%% file: main.tex
\documentclass[letterpaper, 10 pt, journal, twoside]{ieeeconf}

\IEEEoverridecommandlockouts
\usepackage{amsmath, amssymb, cite, graphicx, url, algorithm2e, subfigure, tabularx, booktabs, multirow}

\DeclareMathOperator*{\argmax}{arg\,max}
\DeclareMathOperator*{\argmin}{arg\,min}
\def\eg{{\it e.g.}}

\def\ie{{\it i.e.}}

\usepackage{times}
\usepackage{tcolorbox}
\usepackage{mathtools,physics}
\usepackage{bm}
\usepackage{hyperref}

\usepackage{todonotes}

\title{Benchmarking Actor-Critic Deep Reinforcement Learning Algorithms for Robotics Control with Action Constraints}
\author{Kazumi Kasaura$^{*1}$, Shuwa Miura$^{*2}$, Tadashi Kozuno$^{1}$, Ryo Yonetani$^{1}$, Kenta Hoshino$^{3}$, Yohei Hosoe$^{4}$
\thanks{$^{1}$ KK, TK, and RY are with OMRON SINIC X Corporation, Hongo, Bunkyo-ku, Tokyo, Japan. {\tt\small \{kazumi.kasaura, tadashi.kozuno, ryo.yonetani\}@sinicx.com}.
$^{2}$ SM is with Manning College of Information and Computer Sciences University of Massachusetts Amherst, Amherst, Massachusetts. This work was done while he was a research intern at OMRON SINIC X Corporation. {\tt\small smiura@cs.umass.edu}.
$^{3}$ KH is with Graduate School of Informatics, Kyoto University, Sakyo-ku, Kyoto, Japan. {\tt\small hoshino@i.kyoto-u.ac.jp}.
$^{4}$ YH is with Graduate School of Engineering, Kyoto University, Nishikyo-ku, Kyoto, Japan. {\tt\small hosoe@kuee.kyoto-u.ac.jp}.
$^*$Equal contribution.}
}

\begin{document}

\maketitle

\begin{abstract}%
This study presents a benchmark for evaluating action-constrained reinforcement learning (RL) algorithms.
In action-constrained RL, each action taken by the learning system must comply with certain constraints.
These constraints are crucial for ensuring the feasibility and safety of actions in real-world systems. 
We evaluate existing algorithms and their novel variants across multiple robotics control environments, encompassing multiple action constraint types. Our evaluation provides the first in-depth perspective of the field, revealing surprising insights, including the effectiveness of a straightforward baseline approach. The benchmark problems and associated code utilized in our experiments are made available online at \href{https://github.com/omron-sinicx/action-constrained-RL-benchmark}{github.com/omron-sinicx/action-constrained-RL-benchmark} for further research and development.
\end{abstract}


\section{Introduction}
Action-constrained reinforcement learning (RL) imposes explicit constraints on actions taken by the learning system.
These constraints can come from a variety of sources, such as physical limitations of robots (\eg, torque or power limits) \cite{fujitaClippedActionPolicy2018} and safety considerations.
In manufacturing applications, for example, action constraints can prevent robot arms from hitting obstacles~\cite{phamOptLayerPracticalConstrained2018} or moving beyond designated boundaries~\cite{guDeepReinforcementLearning2017}.
Other examples include collision avoidance for autonomous vehicles~\cite{chengEndtoEndSafeReinforcement2019} and maintenance for energy-efficient building operations~\cite{chenEnforcingPolicyFeasibility2021}.
For real systems,
it is important that each action should satisfy constraints throughout the training process, not just at the end, to guarantee its feasibility \cite{phamOptLayerPracticalConstrained2018}.

Many works have explored how to introduce action constraints to existing deep RL algorithms~\cite{phamOptLayerPracticalConstrained2018,dalalSafeExplorationContinuous2018,bhatiaResourceConstrainedDeep2019,chenEnforcingPolicyFeasibility2021,linEscapingZeroGradient2021}.
Deep RL algorithms, which use neural network policies, have achieved success in various continuous control tasks \cite{lillicrapContinuousControlDeep2015,haarnojaSoftActorCriticOffPolicy2018a}.
However, without explicit constraints, agents can try to execute infeasible actions in the environment.
Existing work has addressed this problem by projecting the outputs of policies onto feasible actions, ensuring that constraints are satisfied before actions are executed. 
Examples of algorithms that adopt this approach encompass differentiable optimization layers~\cite{phamOptLayerPracticalConstrained2018,dalalSafeExplorationContinuous2018,chengEndtoEndSafeReinforcement2019, bhatiaResourceConstrainedDeep2019,chowLyapunovbasedSafePolicy2019a,chenEnforcingPolicyFeasibility2021}, Neural Frank-Wolfe Policy Optimization (NFWPO)~\cite{linEscapingZeroGradient2021}, and $\alpha$-projection \cite{sanketSolvingOnlineThreat2020}. Despite the growing interest in action-constrained RL, a comprehensive comparison of these algorithms has not been conducted.

In this paper, we present the first benchmark study for the existing action-constrained RL algorithms.
Our primary contribution is the evaluation of action-constrained RL algorithms in terms of learning performance and computational time requirements, laying the foundation for future research in this domain.
To this end, our evaluation uses various simulated, and thus easy-to-reproduce, robotics control tasks from MuJoCo \cite{todorov2012mujoco} and PyBullet-Gym ~\cite{benelot2018} on OpenAI gym \cite{OpenAIGym} and evaluate existing work with several different types of action constraints.
Our evaluation centers on off-policy actor-critic deep RL algorithms, such as Deep Deterministic Policy Gradients (DDPG) \cite{lillicrapContinuousControlDeep2015} and Soft Actor-Critic (SAC) \cite{haarnojaSoftActorCriticOffPolicy2018a}, due to their superior sample efficiency.
Actor-critic algorithms \cite{konda1999actor} involve training both an actor and a critic. The actor adjusts policy parameters using the policy gradient, while the critic learns the Q-function.

In addition to evaluating existing action-constrained RL algorithms, we also introduce several variants of these algorithms in our evaluation. One of these variants is a simple method that treats action constraints as part of the state transition of the environment, serving as a baseline for comparison with other approaches to action constraints. We also propose an alternative action mapping method called radial squashing, which can be seen as a natural variant of $\alpha$-projection. In our evaluation, we use a variant of DDPG called Twin Delayed DDPG (TD3)~\cite{fujimotoAddressingFunctionApproximation2018} and SAC as the base deep RL algorithms. DDPG variants are common choices for this type of problem~\cite{bhatiaResourceConstrainedDeep2019,linEscapingZeroGradient2021}, while the use of action constraints with SAC has not been previously explored and requires nontrivial modifications to the original algorithm. We describe how to incorporate action constraints into SAC. 

Our experimental results suggest 
that a simple approach that trains the critic with pre-projected actions is empirically shown to achieve performance comparable to that of specialized algorithms. 
Moreover, the results also show that
alternative mapping techniques ($\alpha$-projection and radial squashing) 
can achieve competitive performance compared to the existing algorithms, while requiring less computation time.
On the other hand, the results show that the use of differentiable optimization layers, a common method for action-constrained RL, does not outperform the simple baseline and can take a long time to run. 

\section{Background}
\label{sec:background}
\subsection{Action-Constrained Reinforcement Learning}
We consider an RL problem, where an agent interacts with its environment modeled by a Markov decision process (MDP).
An MDP is represented by a tuple $M=\langle \mathcal{S}, \mathcal{A}, T, d_0, d_R, \gamma \rangle$, where $\mathcal{S}$ and $\mathcal{A}$ in this work are continuous state and action spaces, respectively.
$T$ is a conditional density function describing the dynamics of the environment ($S_{t+1} \sim T(S_t, A_t)$), and
$d_0$ is an initial state distribution.
$d_R$ describes how rewards are generated ($R_t \sim d_R(S_t, A_t, S_{t+1}))$, and
$\gamma$ is a discount factor.
Given the current state, an agent determines what action to take based on its \emph{policy} parameterized by $\theta$, which is either deterministic (represented by $\mu_\theta$; mapping a state to an action) or stochastic ($\pi_\theta$; mapping the state to the probability distribution on $\mathcal{A}$).
The goal of RL is to find a policy that maximizes the expected discounted return $J(\pi_\theta) = \mathbb{E}[\sum_{t=0}^{\infty} \gamma^t R_t|d_0, \pi_\theta]$.
The \emph{action value function} or \emph{Q-function} for policy $\pi$ is defined as $Q^{\pi}(s, a) = \mathbb{E} [\sum_{t=0}^{\infty} \gamma^t R_t|S_0=s, A_0=a, \pi]$.

In this paper, we consider action-constrained RL problems, where
for each state $s \in S$, there is a feasible set of actions $\mathcal{A}_s \subseteq \mathcal{A}$.
As in most previous work, we assume $\mathcal{A}_s$ to be known apriori and characterized by linear~\cite{phamOptLayerPracticalConstrained2018,bhatiaResourceConstrainedDeep2019} or convex constraints~\cite{linEscapingZeroGradient2021,chenEnforcingPolicyFeasibility2021}.
Note that in action-constrained RL, actions are required to satisfy constraints throughout training.\footnote{The exception can be found in \cite{liAugmentedLagrangianMethod2021}, where constraints are not enforced during training.}
Action constraints can arise from a variety of sources such as
physical limitations of robots~\cite{fujitaClippedActionPolicy2018}.
Such constraints can also be used to guarantee the safety of the system.
For example, several previous studies have used Control Barrier Functions (CBF) \cite{amesControlBarrierFunctions2019} as action constraints \cite{chengEndtoEndSafeReinforcement2019, pereiraSafeOptimalControl2021}.

\subsubsection*{Box Constraints}
Existing deep RL algorithms such as DDPG and SAC, by default, are limited to handle 
a specific type of action constraints called
\emph{box constraints}, which have
the form $a_i \in [-a_i^{\max}, a_i^{\max}]$ for each $i$th dimension of the action space.
Input limits like box constraints are almost ubiquitous in continuous control tasks as used in all control tasks from MuJoCo\cite{todorov2012mujoco} in OpenAI gym~\cite{OpenAIGym}.
Existing implementations of deep RL such as \cite{stable-baselines3} commonly handle such box constraints using a hyperbolic tangent function ($\tanh(x)=\frac{e^x-e^{-x}}{e^x+e^{-x}}$) as the final activation layer of policies, which we refer to as \emph{squashing}.
As the outputs of $\tanh$ range from $-1$ to $1$, we can enforce the box constraints by scaling the outputs by $a_i^{\max}$.
While squashing is simple and effective, it cannot handle more complex constraints such as upper-bounds on the weighted sum of $a_i$. 


\vspace{-10pt}
\subsection{Policy Gradient Algorithms}
Policy gradient algorithms are widely used for continuous control tasks.
These algorithms adjust policy parameters $\theta$ using the gradient estimate of the objective function $J(\pi_{\theta})$. 
For deterministic policies $\mu_{\theta}$, the deterministic policy gradient \cite{silverDeterministicPolicyGradient2014} is given by:
\begin{align}
\textstyle
\nabla_{\theta} J(\mu_{\theta}) &= \mathbb{E}_{s \sim \rho^{\mu_\theta}}[\nabla_{a}Q^{\mu_\theta}(s, a)|_{a=\mu_{\theta}(s)}\nabla_{\theta}\mu_{\theta}(s)],
\end{align}
where $\rho^{\mu_{\theta}}$ is a discounted state distribution under $\mu_{\theta}$.

Deep Deterministic Policy Gradient (DDPG) \cite{lillicrapContinuousControlDeep2015} is a model-free actor-critic algorithm that combines the deterministic policy gradients with function approximation using neural networks.
DDPG uses off-policy data (replay buffer) to train the critic ($Q_w$ where $w$ represents the weights) and uses the critic to train the actor ($\mu_{\theta}$).
For each policy update step, a minibatch of transitions ($\{s_i, a_i, r_i, s'_i\}_{i=1}^N$) is sampled from the replay buffer. 
Then DDPG updates its critic in the following direction:
\begin{equation}
\label{eq:critic}
\textstyle
\nabla_w \frac{1}{N} \sum_{i=1}^N (Q_w(s_i, a_i) - y(s'_i, r_i))^2
\end{equation}
where $y(r_i, s_i) = r_i + \gamma Q_{w'}(s'_i, \mu_{\theta'}(s'_i))$.
The target networks, represented by $Q_{w'}$ and $\mu_{\theta'}$, have distinct parameters $w'$ and $\theta'$ and serve to stabilize the training process.
DDPG next updates its actor using the following estimate for the policy gradient:
\begin{equation}
\textstyle
\label{eq:naive_projection}
\nabla_{\theta} J(\mu_{\theta}) \approx 
\frac{1}{N} \sum_{i=1}^N \nabla_{a}Q_w(s_i, a)|_{a={\mu_{\theta}(s_i)}}\nabla_{\theta} \mu_{\theta}(s_i)
\end{equation}
As described above, DDPG handles box constraints via squashing.

Twin Delayed DDPG (TD3) \cite{fujimotoAddressingFunctionApproximation2018} is a refined version of DDPG that addresses the approximation errors often encountered in DDPG. TD3 introduces several improvements over the original DDPG algorithm. One notable enhancement is the \emph{clipped double-Q} trick, which employs the minimum value of two Q-function outputs to decrease overestimation bias, thereby enhancing the stability and performance of the algorithm.

Soft Actor Critic (SAC) \cite{haarnojaSoftActorCriticOffPolicy2018a} is a model-free actor-critic algorithm that optimizes stochastic policies with the following entropy-regularized objective:
\begin{equation}
\label{eq:sac}
\textstyle
\mathbb{E}[\sum_{t=0}^{\infty} \gamma^t R_t + \alpha H(\pi(\cdot |S_t))|d_0, \pi]
\end{equation}
where $H$ is the entropy and $\alpha > 0$ is an entropy coefficient.
The entropy regularization term in the SAC algorithm encourages exploration during training by promoting a more stochastic policy.
The critic in SAC learns the Q-function incorporating the entropy term (the soft Q-function), which is trained using the clipped double-Q trick as in TD3.
The actor is then updated using:
\begin{equation}
\textstyle
\label{eq:sac}
\nabla_{\theta} \frac{1}{N} \sum_{i=1}^N \min_{j=1,2} Q_{w_j}(s_i, \tilde{a}_{\theta}(s_i)) - \alpha \pi_{\theta}(\tilde{a}_{\theta}(s_i)|s_i)
\end{equation}
where $\tilde{a}_{\theta}(s_i)$ is sampled from a squashed Gaussian policy $\pi_{\theta}(\cdot|s_i)$ using the \emph{reparametrization trick} \cite{haarnojaSoftActorCriticOffPolicy2018a}.



\begin{figure*}[h]
\begin{minipage}{0.75\linewidth}
  \centering
  \subfigure[Closest point projection]{\label{fig:closest_point}\includegraphics[width=0.3\textwidth]{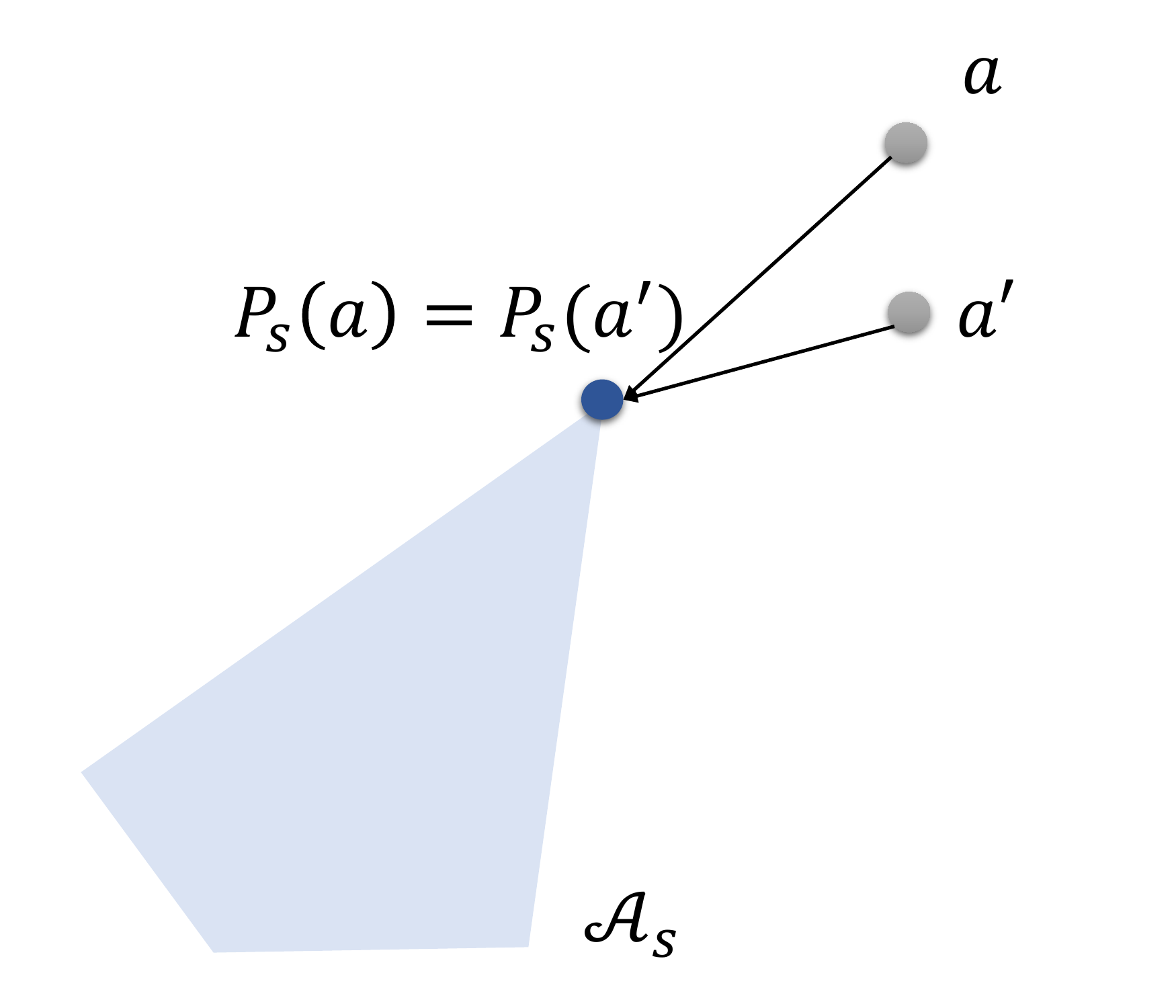}}
  \subfigure[$\alpha$-projection]{\label{fig:alpha-projection}\includegraphics[width=0.3\textwidth]{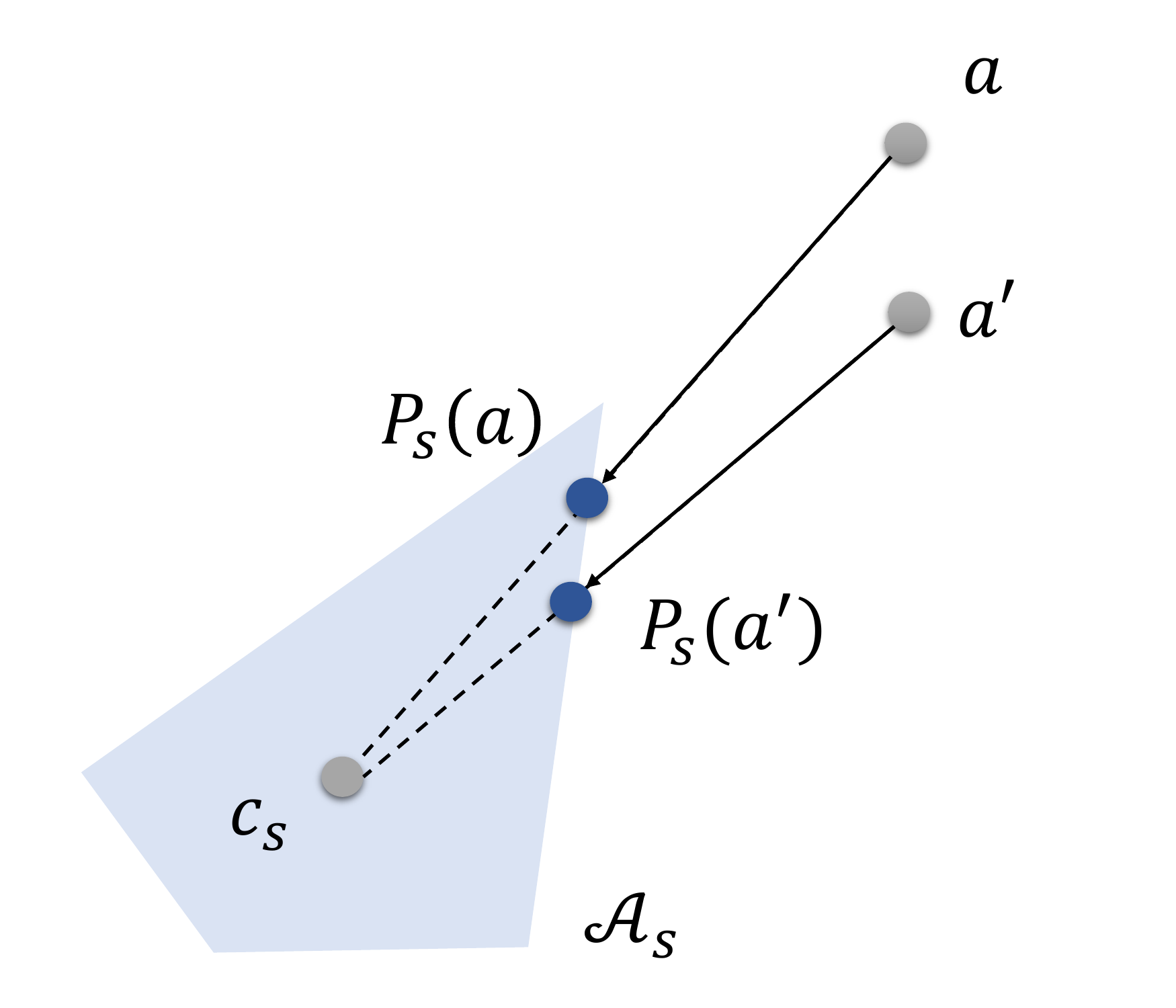}}
  \subfigure[Radial squashing]{\label{fig:radial_squashing}\includegraphics[width=0.3\textwidth]{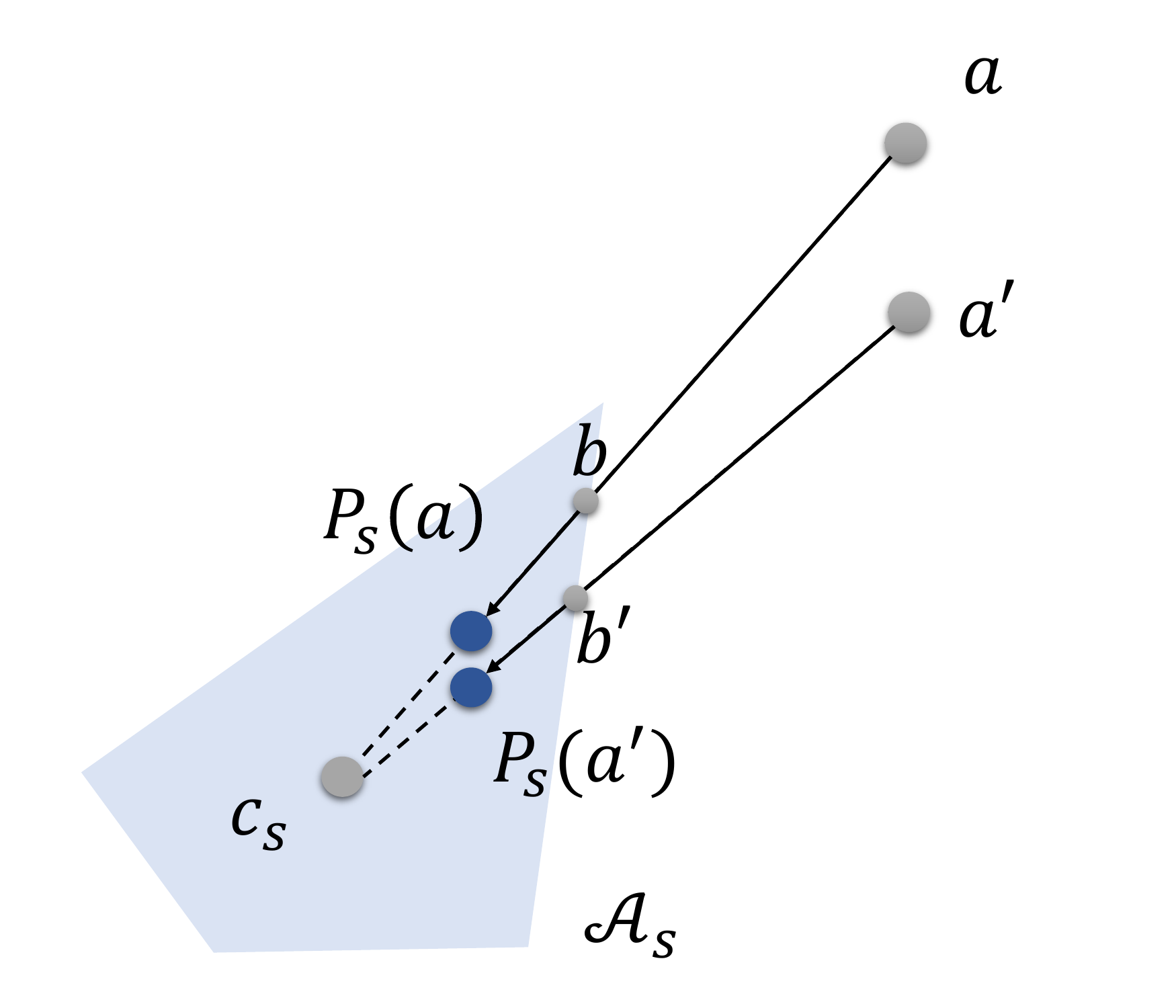}}
  \caption{Illustrations of different mappings used in the paper. $a$ and $a'$ are actions before mappings. $P_s(a)$ and $P_s(a')$ are the corresponding actions after mappings. $\mathcal{A}_s$ is the set of feasible actions. $c_s$ is the Chebyshev center of $\mathcal{A}_s$. }
\end{minipage}
\begin{minipage}{0.24\linewidth}
\includegraphics[width=1.0\textwidth]{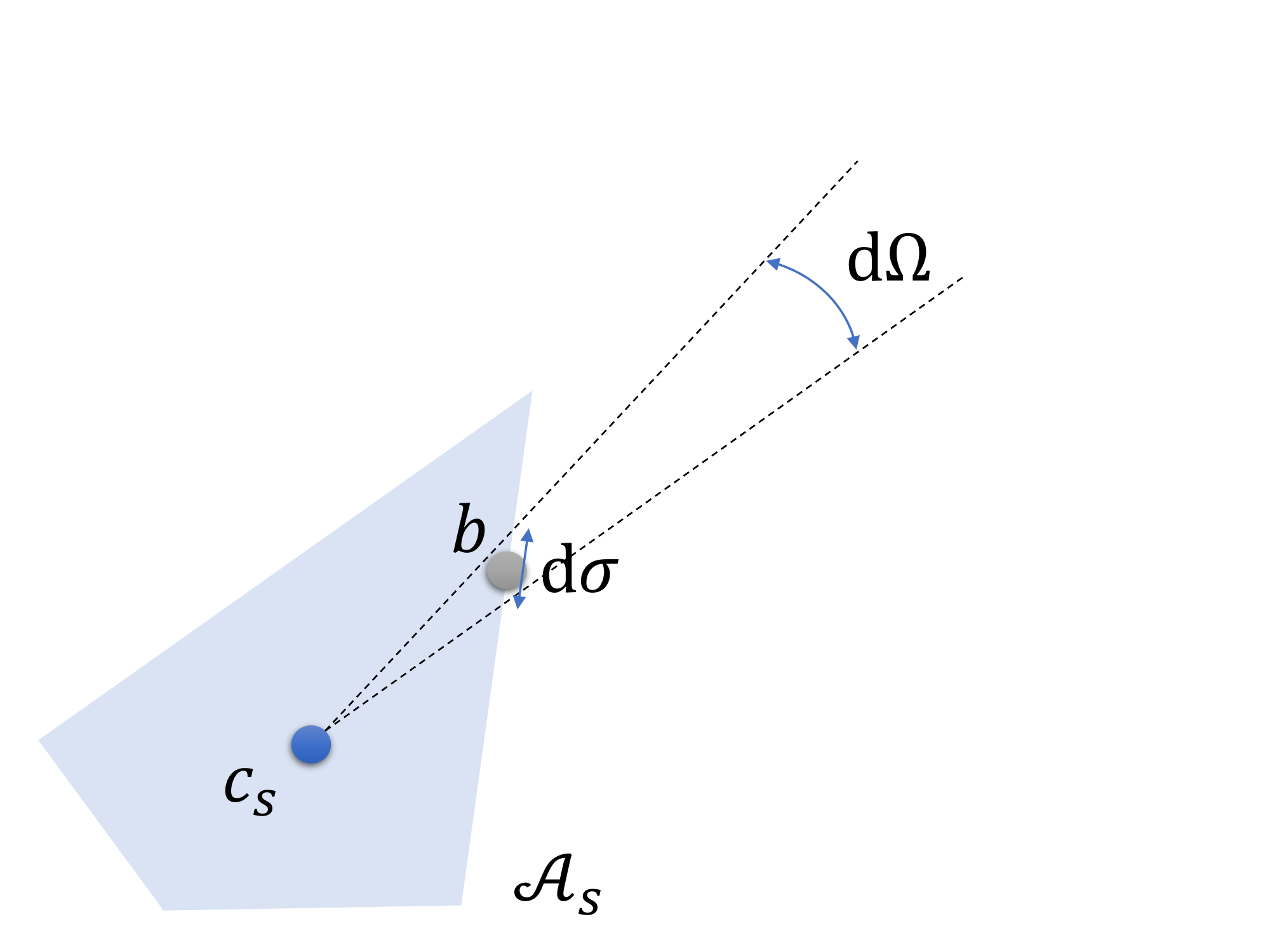}
\caption{Illustration of the volume element $\dd{\sigma}$ of $\partial \mathcal{A}_s$ and the differential solid angle $\dd{\Omega}$ corresponding to it}
\label{fig:alpha-SAC}
\end{minipage}
\end{figure*}

\section{Algorithms for Action-Constrained RL}
In this section, we overview algorithms for action-constrained RL.
To handle non-trivial constraints, most methods 
such as differentiable optimization layers~\cite{phamOptLayerPracticalConstrained2018,dalalSafeExplorationContinuous2018,chengEndtoEndSafeReinforcement2019, bhatiaResourceConstrainedDeep2019,chenEnforcingPolicyFeasibility2021}, NFWPO~\cite{linEscapingZeroGradient2021}, and $\alpha$-projection \cite{sanketSolvingOnlineThreat2020}, 
use a mapping to feasible action sets ($P: \mathcal{S} \times \mathcal{A} \rightarrow \mathcal{A}_s$).
\footnote{Note that mappings discussed here are on actions. This contrasts with projected gradient algorithms, where the 
parameters ($\theta$) are projected to satisfy the constraints.
With neural network policies, projecting $\theta$ so
that $\mu_{\theta}$ satisfies constraints in every continuous state, in general, is not trivial.}
This is because the outputs of policies before the mapping, $\mu_{\theta}(s)$, do not necessarily satisfy action constraints.
Then, rather than the original outputs, the corresponding feasible actions after the mapping, $P(s, \mu_{\theta}(s))$ or also denoted as $P_s(\mu_{\theta}(s))$, are executed in the environment.

In what follows, we first present existing algorithms that use the closest projection to feasible actions (Sec.~\ref{sec:closest_point}).
Then we describe other algorithms that use different mappings in Sec.~\ref{sec:diff}. Unless specified otherwise, our discussion in this section assumes the use of DDPG.

\subsection{Algorithms Based on the Closest Point Projection}
\label{sec:closest_point}
%
In this section, we describe a family of algorithms based on the \emph{closest point projection}, \ie, projecting the outputs of policies to the closest feasible actions typically in terms of the Euclidean norm.
The closest point projection
$P_s$ is defined by the following optimization problem:
\begin{align}
\textstyle
  &P_s(\mu_{\theta}(s)) = \argmin_{a \in \mathcal{A}_s} ||a - \mu_{\theta}(s)||_2 
  .
\end{align}
The problem is quadratic programming for linear constraints and convex programming for convex constraints.
The following algorithms differ in how the actor and critic are updated in the presence of the projection.

\subsubsection{Training Critic with Projected Actions}
\label{sec:training_with_projected}
A baseline algorithm in 
some previous work uses the closest point projection and trains the critic with the projected actions \cite{phamOptLayerPracticalConstrained2018, chengEndtoEndSafeReinforcement2019,bhatiaResourceConstrainedDeep2019, linEscapingZeroGradient2021}.
Namely, it trains the critic $Q_w$ in Eq. (\ref{eq:critic}) with $a_i=P_s(\mu_{\theta}(s_i))$.
The algorithm, however, keeps the actor's update rule unchanged. In the case of DDPG, the actor is updated in the direction of $\nabla_{\theta} Q_w(s_i, \mu_{\theta}(s_i))$ (Eq.~(\ref{eq:naive_projection})). Intuitively, the actor is updated obliviously to the projection.

%
As previous work \cite{bhatiaResourceConstrainedDeep2019} points out, however, updating the actor in $\nabla_{\theta} Q_w(s_i, \mu_{\theta}(s_i))$ is not theoretically principled.
This is because Q-values for pre-projected actions ($Q^{\mu}(s, \mu_{\theta}(s_i))$ are ill-defined
when $\mu_{\theta}(s_i)$ are infeasible.
The approach also has a practical issue. Since the critic is only trained with projected actions ($P_s(\mu_{\theta}(s_i))$), the critic is unlikely to learn action values for pre-projected actions ($\mu_{\theta}(s_i)$).

The problem is that, since a parameterized policy $\mu_{\theta}$ is now combined with the projection $P$, 
policy gradients should be computed for $\nabla_{\theta} J(P_s \circ \mu_{\theta}$) instead of for $\nabla_{\theta} J(\mu_\theta)$, as:
\begin{align}
\textstyle
&\mathbb{E}_{s \sim \rho^{P \circ \mu}}[\nabla_{a}Q^{P \circ \mu}(s, a)|_{a=P_s  (\mu_{\theta}(s))}\nabla_{\theta} P_s \circ \mu_{\theta}(s)]  \\
&= \mathbb{E}_{s \sim \rho^{P \circ \mu}}[\nabla_{a}Q^{P \circ \mu}(s, a)|_{a=P_s(  \mu_{\theta}(s))} \frac{d P_s}{d \mu_{\theta}(s)}\nabla_{\theta} \mu_{\theta}(s)] \label{eq:gradient_with_projection}
\end{align}
where $\frac{d P_s}{d \mu_{\theta}(s)}$ is the Jacobian of the projection $P_s$ with respect to suggested actions.

%

\subsubsection{Differentiable Optimization Layer}
\label{sec:opt_layer}
To compute estimates for the policy gradient with projections in Eq.~(\ref{eq:gradient_with_projection}),
most previous work has used \emph{differentiable optimization layers} as the final layer of the policy network~\cite{phamOptLayerPracticalConstrained2018,dalalSafeExplorationContinuous2018, bhatiaResourceConstrainedDeep2019,chenEnforcingPolicyFeasibility2021}.
OptLayer \cite{phamOptLayerPracticalConstrained2018} combined a differentiable quadratic programming (QP) layer \cite{amosOptNetDifferentiableOptimization2017} with deep RL to handle linear constraints.
PROF \cite{chenEnforcingPolicyFeasibility2021} used 
differentiable convex optimization layers \cite{agrawal2019differentiable} to handle a class of convex programming called disciplined convex programming.
Differentiable optimization layers enable the calculation of projection gradients with respect to suggested actions $\frac{d P_s}{d \mu_{\theta}(s)}$, which changes the actor's update to:
\begin{equation}
\label{eq:actor_update_with_projection}
\textstyle
\frac{1}{N} \sum_{i=1}^N \nabla_{a}Q_w(s_i, a)|_{a={P_s(\mu_{\theta}(s_i))}} \frac{d P_s}{d \mu_{\theta}(s)} \nabla_{\theta} \mu_{\theta}(s_i).
\end{equation}
Note that the actor is now updated in a way that incorporates the effect of the projection, \ie, $\frac{d P_s}{d \mu_{\theta}(s)}$.

Existing methods using differentiable optimization layers share some common challenges. Unlike the original actor update in Eq.~(\ref{eq:naive_projection}), the new update rule in Eq.~(\ref{eq:actor_update_with_projection}) additionally involves projections ($P_s$) and the Jacobian ($\frac{d P_s}{d \mu_{\theta}(s)}$), which incurs non-negligible computational overhead to processing each state in a mini-batch.
More importantly, closest point projection with differentiable optimization layers can suffer from the \emph{zero-gradient issue}.
That is, since many actions initially violating constraints can be projected to the same action, the gradients with respect to $\theta$ tend to vanish when $\mu_{\theta}$ proposes an action violating constraints.
For example, in Fig.~\ref{fig:closest_point}, both $a=\mu_{\theta}(s)$ and 
 $a'=\mu_{\theta}(s)'$ are projected to the same action.
Since the gradient of the projection is zero at $s$, the gradient of the policy $\frac{d P_s}{d \mu_{\theta}(s)} \nabla_{\theta} \mu_{\theta}(s)$ also vanishes at $s$.
As a result, the actor ends up completely wasting the sample.
To alleviate the zero-gradient issue, some methods introduce penalty terms for constraint violations~\cite{phamOptLayerPracticalConstrained2018,chenEnforcingPolicyFeasibility2021}.

\subsubsection{Neural Frank-Wolfe Policy Optimization (NFWPO) }
\label{sec:nfwpo}
To overcome the zero-gradient issue, 
NFWPO \cite{linEscapingZeroGradient2021} proposes to decouple the projection from actor updates.
This is in contrast to previous methods using differentiable optimization layers,
where 
$\frac{d P_s}{d \mu_{\theta}(s)}$ is a part of the actor update in  Eq.~(\ref{eq:actor_update_with_projection}).
For training the critic $Q_w$, NFWPO is identical to DDPG, and uses Eq.(\ref{eq:critic}) with projected actions.
Regarding the actor training, NFWPO initially determines a reference action $a_s \in \mathcal{A}_s$ for each state $s \in S$ in a mini-batch via the Frank-Wolfe algorithm~\cite{frankAlgorithmQuadraticProgramming1956}:
\begin{equation}
\label{eq:reference_action}
\textstyle
a_s = P_s(\mu_{\theta}(s) + \alpha (c_s - P_s(\mu_{\theta}(s))))
\end{equation}
Here, $\alpha$ represents the Frank-Wolfe learning rate, and
\begin{equation}
\label{eq:frank_wolfe_direction}
\textstyle
c_s = \argmax_{c \in \mathcal{A}_s} \langle c, \nabla_a Q_w(s, a)|_{a=P_s(\mu_{\theta}(s))}\rangle
\end{equation}

Subsequently, the policy parameters $\theta$ are updated to minimize the distance between $\mu_{\theta}(s)$ and the reference action $a_s$.
By doing so, NFWPO avoids computing $\frac{d P_s}{d \mu_{\theta}(s)}$ and policy gradients, thereby overcoming the zero-gradient issue.
Without any action constraints, NFWPO is equivalent to DDPG \cite{linEscapingZeroGradient2021}.

%
%

\subsubsection{Training Critic with Pre-Projected Actions}\label{subsec: EnvWrapper}
In addition to the existing methods above, we consider another simple, yet surprisingly effective baseline that trains the actor and the critic with \emph{pre-projected}, possibly infeasible actions $\mu_\theta(s)$, while the projection of the actions to feasible ones is done as a part of state transitions of the environment. 
In this method, the critic learns $\tilde{Q}^{\mu} = Q^{\mu} \circ P$ instead of $Q^{\mu}$, as it is defined even for infeasible actions.
In other words, the critic now learns action-values for suggested actions, given that suggested actions might be projected to feasible actions.
Then, the actor is updated using Eq.~(\ref{eq:naive_projection}).
This approach is computationally more efficient than other techniques using differentiable optimization or NFWPO, since the projection is performed only during rollouts.
Moreover, this approach can be easily combined with other RL algorithms, including SAC.

%


Note that this baseline has several known limitations. As actions are projected to feasible ones as part of state transitions, agents must learn the effects of projection through experienced transitions and rewards. Moreover, this approach can suffer from the zero-gradient issue and may require additional penalty terms to alleviate the problem.


\vspace{-10pt}
\subsection{Other Mapping Techniques}
\label{sec:diff}
In this section, we present several other differentiable mapping techniques. 
Similar to using differentiable optimization layers (Sec~\ref{sec:opt_layer}), the algorithms in this section employ differentiable mappings to feasible actions as the final layers of policies. However, the techniques presented here do not necessarily map actions to the closest feasible ones. The key insight is that since RL algorithms can learn to adjust their suggestions to achieve better performance, there is no inherent need for using the closest point projection.

\subsubsection{$\alpha$-Projection Layer}
\label{sec:alpha}
$\alpha$-projection \cite{sanketSolvingOnlineThreat2020} can be used as an alternative to the closet point projection.
$\alpha$-projection assigns an interior point $c_s \in \mathrm{Int}(\mathcal{A}_s)$ for each state $s \in \mathcal{S}$.
As Fig.~\ref{fig:alpha-projection} shows, given a suggested action $a \in \mathcal{A}$, $\alpha$-projection moves $a$ toward $c_s$ until it reaches $\mathcal{A}_s$.
Formally, it picks an action on the ray from $c_s$ to $a$ that is closest to $a$ (each $a \in \mathcal{A}_s$ is mapped to itself).
Since 
points on different rays are mapped to different points,
the gradient does not vanish in the example.

As originally studied in \cite{sanketSolvingOnlineThreat2020}, for linear-inequality constraints, a closed-form expression exists for $\alpha$-projection and can be implemented as an additional neural network layer. 
The actor is then updated using Eq.~(\ref{eq:actor_update_with_projection}), where $P$ is replaced with $\alpha$-projection.
It is also possible to extend the $\alpha$-projection to problems with an elliptical constraint, which has the form $(a-c(s))^{\mathsf{T}}Q(s)(a-c(s))-b(s)$ where $Q(s)$ is a positive-semi-definite matrix. 

Note that $\alpha$-projection assumes the existence of an interior point $c_s \in \mathrm{Int}(\mathcal{A}_s)$ such that, for any point $x \in \mathcal{A}_s$, the segment between $c_s$ and $x$ is contained in $\mathcal{A}_s$. 
For linear constraints, we use the Chebychev center of $\mathcal{A}_s$ as $c_s$ as in \cite{sanketSolvingOnlineThreat2020}. 
%
During training of the actor, $\alpha$-projection must be performed for each state in a minibatch. Computing $c_s$ for each state can be time-consuming, so we store the previously computed Chebyshev centers in the replay buffer and reuse them during training.
For problems with an elliptical constraint, we use $c(s)$ as $c_s$.

\subsubsection{Radial Squashing Layer}
\label{sec:radial}
We also propose an alternative mapping to feasible actions called \emph{radial squashing}, which can also be implemented as a neural network layer.
Instead of clipping the ray from $c_s$ as in $\alpha$-projection,
radial squashing shrinks actions into the feasible set of actions around its center $c_s$, as illustrated in Fig.~\ref{fig:radial_squashing}.
 Let $a$ be the given point and $b$ be the intersection of the boundary of $\mathcal{A}_s$ and the ray from $c_s$ to $a$. The radial squashing layer maps:
\begin{equation}\label{eq:radial squashing}
a\mapsto c_s + \tanh\left(\|a-c_s\|/\|b-c_s\|\right)(b-c_s).
\end{equation}
Note that, this mapping is differentiable at $c_s$ (the Jacobian becomes the identity matrix there) and
gradients never vanish for any direction.

\vspace{-5pt}
\subsection{Incorporating Action Constraints to SAC}
\label{sec:sac}
Although SAC has been a popular algorithm for general RL problems and can be easily combined with the approach described in \ref{subsec: EnvWrapper}, combining it with mapping techniques for action-constrained RL is not trivial. This is mainly due to the entropy term in Eq.~(\ref{eq:sac}) that requires the calculation of the probability density of a policy after actions are mapped to feasible ones.
In general, for a random variable $X$ with a probability density function $p$, the probability density function $q$ of $f(X)$ after a differentiable mapping $f$ is given by:
\begin{equation}
    q(f(X))=|\det J_f(X)|^{-1}p(X),
\end{equation}
when $\det J_f(X) \neq 0$, where $J_f(X)$ is the Jacobian of $f$.

\subsubsection{Radial Squashing with SAC}
\label{sec:sacr}
Combining radial squashing with SAC is straightforward since the determinant of Jacobian never vanishes.
To evaluate the change in probability density by the radial squashing, it is enough to calculate the Jacobian of Eq.~(\ref{eq:radial squashing}) as follows: 
\begin{equation}\label{eq:Jacobian of radial squashing}
(a-c_s)\left(\mathrm{grad}\, \frac{\tanh L}{L}\right)^{\mathsf{T}}+\frac{\tanh L}{L}I,
\end{equation}
where $I$ is the identity matrix and $L:= \|a-c_s\|/\|b-c_s\|$ is a function of $a$ that depends on the constraint.

\subsubsection{$\alpha$-projection with SAC}
\label{sec:saca}
On the other hand, combining the $\alpha$-projection layer with SAC is non-trivial because the probability density after projection may become infinity (\ie, the determinant of Jacobian may be zero.) To overcome this difficulty, we consider two probability density functions for the policy after projection: a $d$-dimensional density function on the interior of $\mathcal{A}_s$ and a $(d-1)$-dimensional density function on the boundary of $\mathcal{A}_s$, where $d$ is the dimension of $\mathcal{A}_s$. The entropy of the distribution is defined by the sum of the entropies for these distributions on the interior and boundary. Mathematically, we define the measure of $\mathcal{A}_s$ as the sum\footnote{The weight of the entropy (or the measure) for the boundary in this sum is arbitrary and not scale-invariant. It can be considered as a hyperparameter of this algorithm. In this paper, we simply take one. Note that the action space is scaled in the implementation of these algorithms.} of the standard measure of the interior of $\mathcal{A}_s$ and that of the boundary of $\mathcal{A}_s$ as a Riemannian submanifold of the Euclidean space~\cite{lee2012smooth}.

While the probability density for the interior of $\mathcal{A}_s$ remains unchanged by the projection, we need to integrate the probability density before projection for the boundary of $\mathcal{A}_s$.
Let $p$ be the probability density function before projection, and let $q$ be the desired probability density function on the boundary after projection. We use the spherical coordinate system with origin $c_s$. Let $b \in \partial \mathcal{A}_s$ be the point on the boundary, and let $\dd{\sigma}$ be the volume element of $\partial \mathcal{A}_s$ at $b$. We must integrate $p$ on the solid cone projected to $\dd{\sigma}$:
\begin{equation}
    q \dd{\sigma} = \int_{r_0}^{\infty} pr^{d-1}\dd{r}\dd{\Omega},
\end{equation}
\looseness=-1
where $r_0:=\|b-c_s\|$ and $\dd{\Omega}$ is the differential solid angle corresponding to $\dd{\sigma}$ (see Fig.~\ref{fig:alpha-SAC}). This can be calculated by
    $\dd{\Omega} = r_0^{-1}\cos \theta \dd{\sigma}$,
where $\theta$ is the angle between the vector $b-c_s$ and the normal vector of $\partial \mathcal{A}_s$ at $b$.
Since $p$ is a Gaussian distribution for SAC, it is enough to calculate an integration of form $\int r^{d-1}\exp(-(Ar+B)^2)\dd{r}$ after completing the square. We can get its explicit expression using $\exp(-(Ar+B)^2)$ and $\mathrm{erf}(Ar+B)$, where $\mathrm{erf}$ is the error function, by applying partial integration repeatedly.



\subsubsection{Remark on Differential Optimization Layer}
\label{sec:remark_differential}
Combining differentiable optimization layers with SAC, on the other hand, may not be feasible. 
This is because closest point projection can map 
a region with more than one dimension to one point, making it difficult to
determine the change in
probability distribution. 

%
\vspace{-10pt}
\subsection{ConstraintNet}
Another possible approach for handling action-constraints is to modify the output layer of the actor network based on the type of constraints. ConstraintNet~\cite{brosowskySampleSpecificOutputConstraints2021} uses a convex combination of $n$ given vertices as the output layer when the feasible set of actions is a convex polytope and the vertices of the polytope are known. However, this approach has some limitations: the number of vertices in a polytope can be exponential in the dimension of the action, and it is not straightforward to extend this approach to state-dependent constraints. Therefore, we did not include this approach in our evaluations.


\section{Experiments}
\label{sec:experiments}
In this section, we conduct an empirical evaluation of action-constrained RL algorithms on various simulated control tasks from PyBullet-Gym~\cite{benelot2018} and MuJoCo~\cite{todorov2012mujoco} in OpenAI Gym~\cite{OpenAIGym}. We additionally assess the running time of each algorithm in Sec.~\ref{sec:exp3}.

\renewcommand{\labelitemii}{$\circ$}
\vspace{-10pt}
\subsection{Algorithms}
We compare the following 13 algorithms in total.
\begin{itemize}
\item TD3 family:
\begin{itemize}
    \item \textbf{DPro}: TD3 with critic trained using projected actions (Sec.~\ref{sec:training_with_projected})
    \item \textbf{DPro+}: \textbf{DPro} with the penalty term (see below)
    \item \textbf{DPre}: TD3 with pre-projected actions (Sec.~\ref{subsec: EnvWrapper})
    \item \textbf{DPre+}: \textbf{DPre} with penalty term
    \item \textbf{DOpt}: TD3 with optimization layer (Sec.~\ref{sec:opt_layer})
    \item \textbf{DOpt+}: \textbf{DOpt} with penalty term
    \item \textbf{NFW}: NFWPO (Sec.~\ref{sec:nfwpo}) with TD3 techniques (clipped double Q learning, target policy smoothing and delayed policy update)
    \item \textbf{DAlpha}: TD3 with $\alpha$-projection (Sec.~\ref{sec:alpha})
    \item \textbf{DRad}: TD3 with radial squashing (Sec.~\ref{sec:radial})
    \end{itemize}
    \item SAC family:
    \begin{itemize}
    \setcounter{enumi}{10}
    \item \textbf{SPre}: SAC with pre-projected actions (Sec.~\ref{subsec: EnvWrapper})
    \item \textbf{SPre+}: \textbf{SPre} with penalty term
    \item \textbf{SAlpha}: SAC with $\alpha$-projection (Sec.~\ref{sec:saca})
    \item \textbf{SRad}: SAC with radial squashing (Sec.~\ref{sec:sacr})
    \end{itemize}
\end{itemize}
\textbf{DPro+}, \textbf{DPre+}, \textbf{DOpt+}, and \textbf{SPre+} introduce the penalty term for constraint violations discussed in Sec.~\ref{sec:opt_layer} or Sec.~\ref{subsec: EnvWrapper}. Specifically, we added $\|\max\{Ax-b,\mathbf{0}\}\|$ for linear constraints $Ax\leq b$, where $A$ is normalized so that the norm of each row is $1$, as in \cite{phamOptLayerPracticalConstrained2018}. For elliptical constraint $(x-c)^{\mathsf{T}}Q(x-c)\leq b$, $\max\{\sqrt{(x-c)^{\mathsf{T}}Q(x-c)}-\sqrt{b},0\}$ was added where $Q$ is normalized so that $\tr Q$ equals to the dimension. In case there exist both constraints, we take the sum of two penalties.

\begin{table}
\begin{center}
\scalebox{0.9}{
\begin{tabular}{ lcccccc } 
\\
 \toprule
 
Hyperparameters & \multicolumn{2}{c}{Reacher} &\multicolumn{2}{c}{Hopper}&\multicolumn{2}{c}{Others} \\ 
& TD3&SAC&TD3&SAC&TD3&SAC\\ 
 \midrule
 Discount factor &  \multicolumn{2}{c}{0.98}&\multicolumn{2}{c}{0.99}&\multicolumn{2}{c}{0.99}\\
 Net. arch. 1st & 400 & 400& 400& 256& 400& 256\\
 Net. arch. 2nd & 300 & 300& 300& 256& 300& 256\\
 Batch size & 100&256&256&256&100&256\\
 Learning rate &1e-3&7.3e-4&3e-4&3e-4&1e-3&3e-4\\
 Buffer size & 2e5&3e5&1e6&1e6&1e6&1e6\\
 Target Update Ratio & 0.005 & 0.02& 0.005 & 0.005 & 0.005 & 0.005 \\
 Action noise & 0.1 & - & 0.1 & - & 0.1 &-\\
 FW learning rate &0.05&-&0.01&-&0.01&-\\
Learning starts & 1e5& 1e5& 1e5& 1e5& 1e5& 1e5\\ Use SDE &-&True&-&False&-&False\\
Train freq.&1 epi.&8 steps&1 step&1 step&1 epi.&1 step\\
Gradient steps& -1 & 8 & 1 & 1&-1&1\\
 \bottomrule
\end{tabular}
}
\end{center}
\caption{Hyperparameters used in the experiments.}
\label{table:hyperparameters}
\vspace{-5pt} 
\end{table}


\vspace{-10pt}
\subsection{Implementation Details}
We adapted the implementations of TD3 and SAC in Stable Baselines 3 (SB3)~\cite{stable-baselines3}, a popular reinforcement learning library using PyTorch~\cite{paszke2019pytorch}. We use tuned hyperparameters in RL Baselines Zoo~\cite{rl-zoo3} for each base algorithm (TD3 or SAC) and each environment. For \textbf{NFW}, the Frank-Wolfe learning rate is tuned to $0.05$ in Reacher and $0.01$ in other environments. Table \ref{table:hyperparameters} shows the used hyperparameters. For train frequency, `1 epi.' means that the model is updated every episode. 
Similarly, `1 step' means that the model is updated every step.
For gradient steps, `-1' means to do as many gradient steps as steps done in the environment during the rollout.

We also used gurobi~\cite{bixby2007gurobi} for solving linear or quadratic programming and cvxpylayers~\cite{agrawal2019differentiable} for differentiable optimization layers.
To address potential failures of the optimization solver for projections from distant points, we applied the squashing technique to each coordinate of the neural network outputs before projection, as done in \cite{linEscapingZeroGradient2021}, for methods utilizing pre-projected actions, NFWPO and the optimization layer.

\vspace{-10pt}
\subsection{Evaluations with Various Environments and Constraints}
\label{sec:exp1}

\begin{figure*}[h]
\begin{minipage}{0.49\linewidth}
    \centering
    \includegraphics[width=\textwidth]{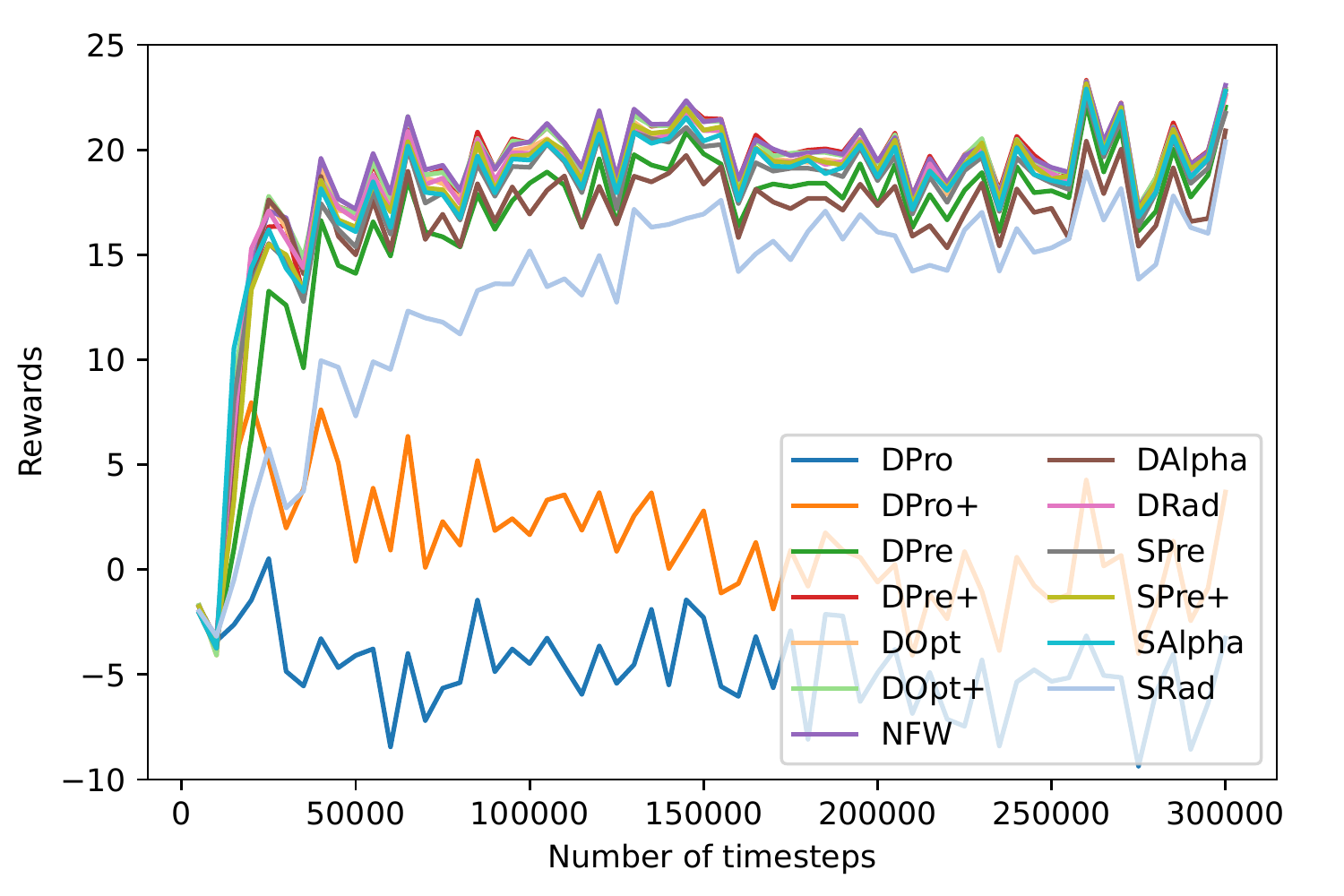}
    \caption{Learning curves for R+L2}\label{fig:reacher}
\end{minipage}
\begin{minipage}{0.49\linewidth}
    \centering
    \includegraphics[width=\textwidth]{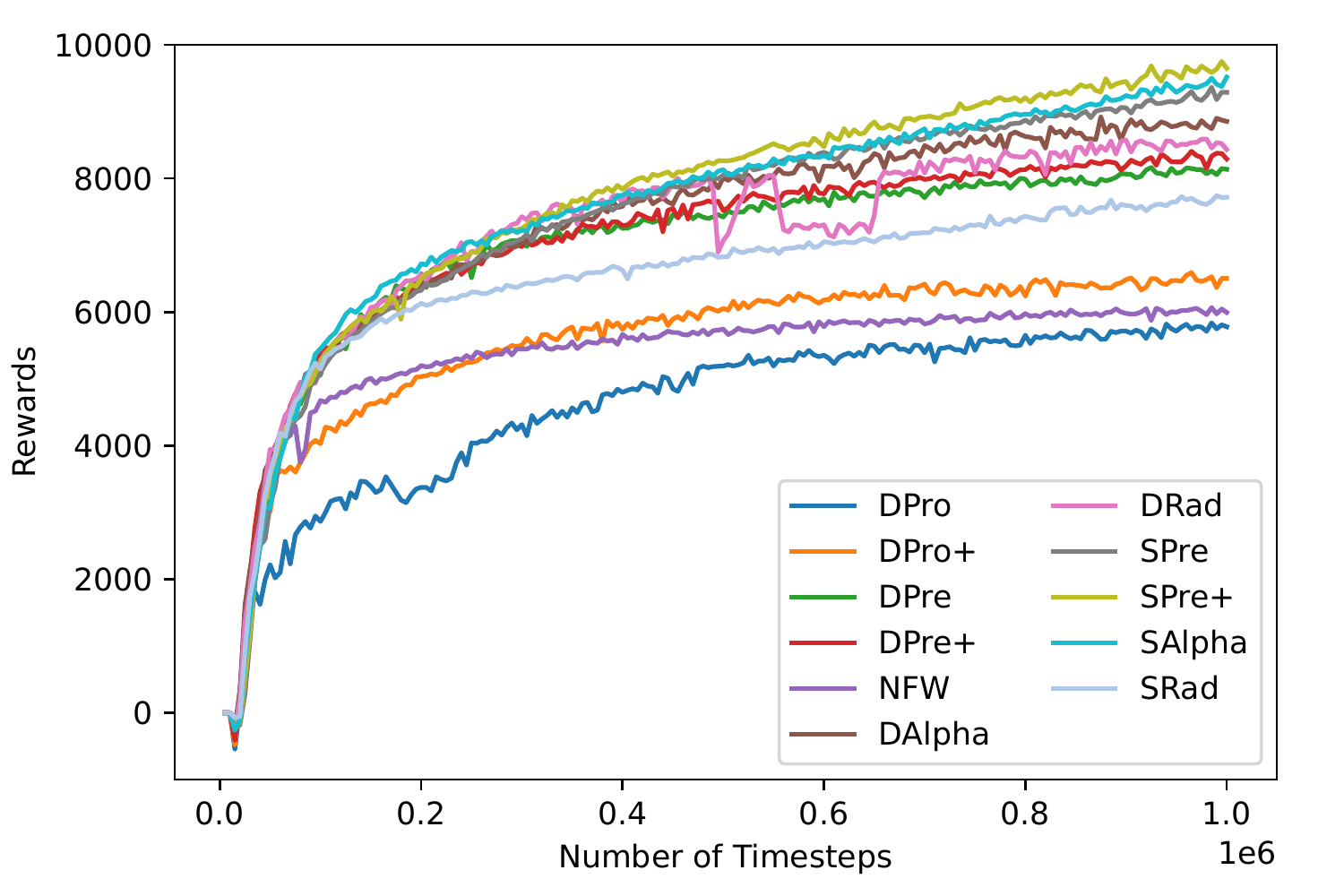}
    \caption{Learning curves for HC+O with batch size $\geq 100$}\label{fig:half_cheetah}
\end{minipage}
\end{figure*}

\begin{table}[t]
    \centering
    \begin{tabular}{lcc}
    \\
    \toprule[1pt]
    Environment & Name & Constraint\\
    \midrule
    Reacher & R+N &  No additional constraint\\
    \cmidrule{2-3}
&R+L2 &$a_1^2+a_2^2\leq 0.05$\\
    \cmidrule{2-3}
&R+O03 &$\sum_{i=1}^2 |w_ia_i|\leq 0.3$\\
    \cmidrule{2-3}
&R+O10 &$\sum_{i=1}^2 |w_ia_i|\leq 1.0$\\
    \cmidrule{2-3}
&R+O30 &$\sum_{i=1}^2 |w_ia_i|\leq 3.0$\\
&R+M &$\sum_{i=1}^2 \max\{w_ia_i,0\}\leq 1.0$\\
    \cmidrule{2-3}
&R+T &$a_1^2+2a_1(a_1+a_2)\cos \theta_2$\\
& &$+(a_1+a_2)^2\leq 0.05$\\
    \midrule

HalfCheetah& HC+O & $\sum_{i=1}^6|w_ia_i|\leq 20$\\
    \cmidrule{2-3}
&HC+MA & $w_1a_1\sin (\theta_1+\theta_2+\theta_3)$\\
&&$+w_4a_4\sin (\theta_4+\theta_5+\theta_6)\leq 5$\\
    \midrule
Hopper&H+M& $\sum_{i=1}^3\max\{w_ia_i,0\}\leq 10$\\
    \cmidrule{2-3}
&H+O+S& $\sum_{i=1}^3|w_ia_i|\leq 10$, $\sum_{i=1}^3 a_i^2\sin^2\theta_i\leq 0.1$\\
    \midrule
Walker2d&W+M & $\sum_{i=1}^6\max\{w_ia_i,0\}\leq 10$\\
    \cmidrule{2-3}
&W+O+S& $\sum_{i=1}^6|w_ia_i|\leq 10$, $\sum_{i=1}^6 a_i^2\sin^2\theta_i\leq 0.1$\\
    \bottomrule
    \end{tabular}
    \caption{Experiment environments and constraints}
    \label{tab:experiment settings}
\vspace{-5pt} 
\end{table}
    
In this experiment, we compared algorithms on the Reacher in PyBullet-Gym and Hopper, Walker2d, and HalfCheetah in MuJoCo\footnote{The descriptions of the tasks are available at www.gymlibrary.dev.} with various constraints.
Each action is represented by a vector $(a_1,\ldots,a_d)$ corresponding to torques given to $d$ joints, where $d=2$ for Reacher, $d=3$ for Hopper and $d=6$ for Walker2d and HalfCheetah. Let $\theta_1,\ldots,\theta_d$ and $w_1,\ldots,w_d$ be the angle and the angular velocity of joints respectively.

In addition to the original box constraint $-1\leq a_i \leq 1$ for each $i$, we consider constraints shown in Table~\ref{tab:experiment settings}.
Note that the constraint for R+L2 and HC+O are the same as those in \cite{linEscapingZeroGradient2021}.
For R+N, R+L2, R+O03, R+O10, R+O30, R+T, HC+O, H+O+S and W+O+S, the center ($c_s$) of the feasible actions is at the origin, but that is not necessarily for other cases. 

For each combination of algorithms and constraints, we ran the algorithm with $10$ different seeds. The number of timesteps is $3\times 10^5$ in Reacher and $10^6$ in other three environments as in \cite{rl-zoo3}. In each run, we evaluated five episodes and took their mean in every $5,000$ timesteps. After each run, we evaluated $50$ episodes again for the best parameters and consider their mean as the final result.

\newcolumntype{C}{>{\centering\arraybackslash}X}
\input{reward_table2.tex}

Table \ref{tab:rewards} shows the average values and standard errors of rewards over the $10$ seeds, where the top three algorithms for each constraint are emphasized with bold text. 
Due to the limitation of cvxpylayer, \textbf{DOpt} and \textbf{DOpt+} cannot be applied to some constraints.\footnote{Constraints should adhere to the disciplined parameterized programming rules~\cite{agrawal2019differentiable}.} So we excluded these algorithms from some results and display ``unavailable" in the table. Furthermore, since \textbf{DOpt} and \textbf{DOpt+} are too computationally expensive to run with the tuned batch size for MuJoCo environments, we excluded them and additionally ran all algorithms including them with batch size $16$ for constraints for which \textbf{DOpt} and \textbf{DOpt+} are available. The bottom three rows of the table shows the results of additional experiments.
Fig.~\ref{fig:reacher} and Fig.~\ref{fig:half_cheetah} present the learning curves averaged over the $10$ seeds for R+L2 and HC+O, respectively.

\vspace{-10pt}
\subsection{Measurements of Training Runtime}
\label{sec:exp3}
\begin{figure}[t]
  \centering
  \includegraphics[width=.8\linewidth]{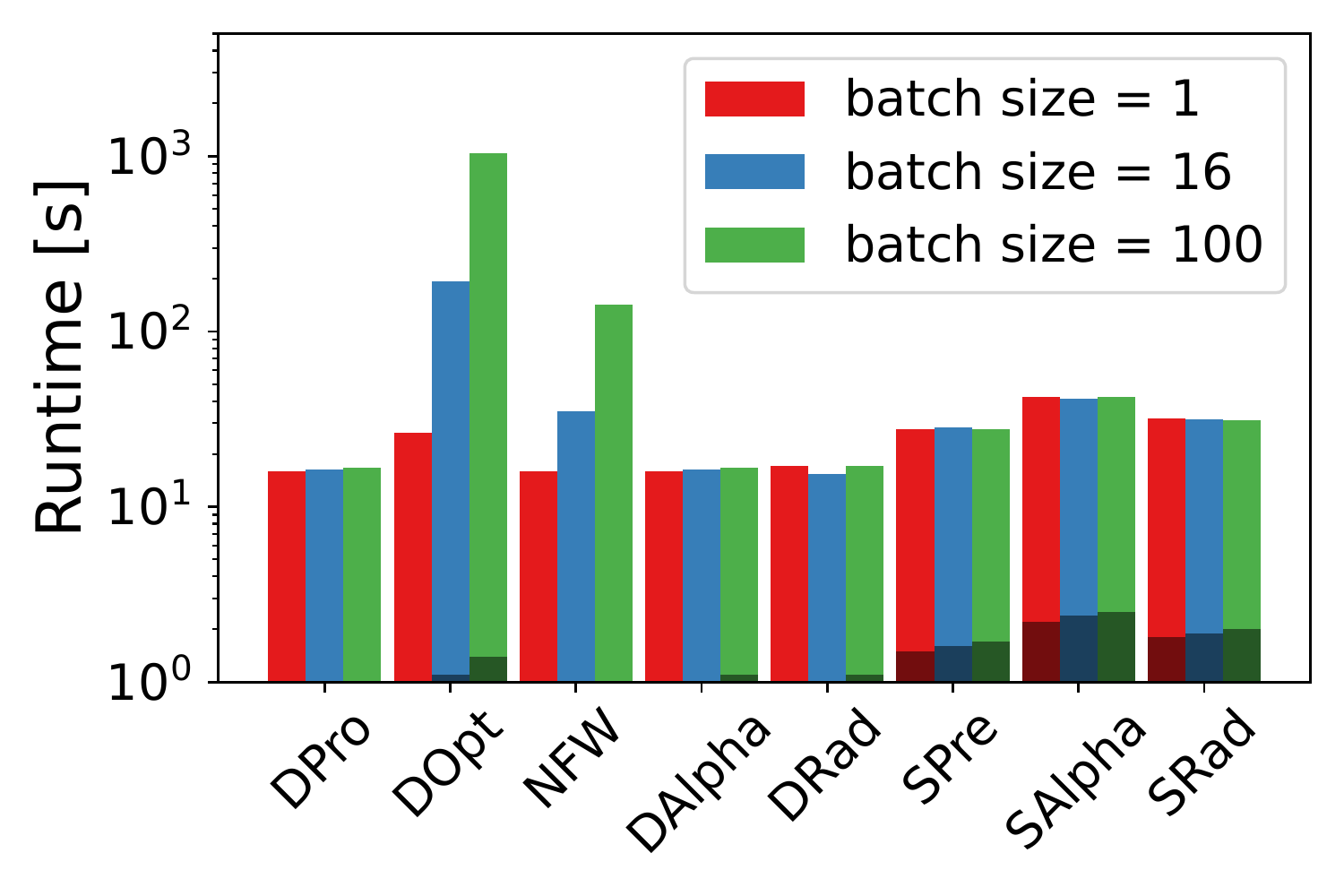}
  \caption{Training runtime for HalfCheetah. The light areas of the bars show the time on a CPU and the dark areas show the time on a GPU.}
  \label{fig:runtime}
\end{figure}

Finally, we measured the runtime of each algorithm for the first $1000$ gradient steps of the training procedure in the HalfCheetah environment, with batch sizes $1, 16$ or $100$.
Fig.~\ref{fig:runtime} summarizes the runtimes in seconds averaged over 10 trials with different seeds.

\vspace{-10pt}
\subsection{Findings and Discussion}

\subsubsection{Training the critic using pre-projected actions is a good baseline especially with penalty terms}
We found that \textbf{DPro}, \ie, training the critic using projected actions, demonstrated limited performance compared to \textbf{DPre} with pre-projected actions in HalfCheetah and Walker environments, despite that it was used as a baseline in some previous work~\cite{bhatiaResourceConstrainedDeep2019,linEscapingZeroGradient2021}. This could be due to its fundamental defect discussed in Section~\ref{sec:training_with_projected} and indicates that \emph{pre-projected actions should be used for training the critic}. 
\textbf{DPre+}, on the other hand, achieved performance comparable to other TD3 variants.
We contend that both \textbf{SPre} and \textbf{SPre+}, which are SAC algorithms utilizing pre-projected actions, serve as viable initial selections among the evaluated algorithms. As demonstrated in Table~\ref{tab:rewards}, these methods consistently rank within the top three across most conditions, while also offering the advantages of relatively low implementation efforts and computational runtime as shown in Fig.~\ref{fig:runtime}

\subsubsection{The use of optimization layers and NFWPO comes with significant runtime overheads}
\textbf{DOpt} requires significant computation time, as clearly shown in Fig.~\ref{fig:runtime}. Nevertheless, it does not demonstrate significantly better performance than \textbf{DPre+}, \textbf{DAlpha} or \textbf{DRad}. 

While \textbf{NFW} achieved the best learning performance among the TD3 family in the Reacher environment, its performance is not significantly better in other environments. On the other hand, it also requires a considerably large runtime as the batch size becomes larger. 


\subsubsection{Mapping techniques can be less expensive alternatives to optimization layers}
\textbf{SAlpha} and \textbf{SRad} demonstrated good performance for a small batch size at the cost of 140-150\% runtime overheads compared to \textbf{SPre+}.
Also, \textbf{DAlpha} and \textbf{DRad} performed on par (if not better) with \textbf{DOpt} and \textbf{DOpt+}, while requiring much less runtime.
Nonetheless, under certain conditions (R+L2 and R+T), SRad exhibited subpar performance. Investigating the reasons behind this underperformance will be the subject of future research.


%
\section{Conclusion}
In this paper, we compared variants of TD3 and SAC 
on a variety of continuous control tasks in the presence of action constraints.
Our evaluation includes 
new variants of the existing action-constrained RL algorithms.
Our benchmark evaluation has led to the following main findings. 1) Training the critic with pre-projected actions is a good baseline, especially with penalty terms. 2) The use of optimization layers and NFWPO comes with significant runtime overheads. 3) Mapping techniques are useful alternatives to optimization layers. A complete implementation of our benchmark evaluation is available online at \href{https://github.com/omron-sinicx/action-constrained-RL-benchmark}{github.com/omron-sinicx/action-constrained-RL-benchmark}. 


\bibliographystyle{IEEEtran}
\bibliography{main}

\end{document}

%% file: reward_table2.tex
\begin{table*}[t]
    \centering
    \begin{tabularx}{\textwidth}{lCCCCCCCCCCCCC}
    \toprule[1pt]
  & \multicolumn{9}{c}{TD3 Family} & \multicolumn{4}{c}{SAC Family} \\ 
 \cmidrule(lr){2-10} \cmidrule(lr){11-14}
 Conditions  & DPro & DPro+ & DPre & DPre+ & DOpt & DOpt+ & NFW & DAlpha & DRad & SPre & SPre+ & SAlpha & SRad \\
    \midrule
R+N &17.57&\multicolumn{5}{c}{Same as DPro due to no constraints}&\textbf{18.47}&17.40&17.34&\textbf{18.61}&Same&15.01&\textbf{18.47}\\
&$\pm 0.27$& & & & & & $\pm 0.22$&$\pm 0.38$&$\pm 0.24$&$\pm 0.23$& SPre &$\pm 2.00$&$\pm 0.26$\\
    \midrule
R+L2 &-4.65&9.90&17.81&\textbf{19.70}&19.12&19.15&\textbf{19.70}&18.46&18.76&18.84&\textbf{19.45}&18.97&15.69\\
&$\pm 0.59$&$\pm 2.69$&$\pm 0.33$&$\pm 0.30$&$\pm 0.30$&$\pm 0.51$&$\pm 0.33$&$\pm 0.86$&$\pm 0.37$&$\pm 0.27$&$\pm 0.29$&$\pm 0.31$&$\pm 0.46$\\
    \midrule
         R+O03 &17.32&18.74&18.17&18.19&17.72&18.55&\textbf{18.86}&18.20&17.88&\textbf{18.78}&15.82&18.58&\textbf{19.11}\\
&$\pm 0.37$&$\pm 0.32$&$\pm 0.30$&$\pm 0.33$&$\pm 0.42$&$\pm 0.26$&$\pm 0.38$&$\pm 0.29$&$\pm 0.32$&$\pm 0.35$&$\pm 0.43$&$\pm 0.27$&$\pm 0.30$\\
    \midrule
         R+O10 &17.44&17.78&17.32&17.76&18.01&17.80&\textbf{18.70}&17.52&17.98&18.30&\textbf{18.37}&18.29&\textbf{18.82}\\
&$\pm 0.53$&$\pm 0.34$&$\pm 0.31$&$\pm 0.34$&$\pm 0.18$&$\pm 0.42$&$\pm 0.30$&$\pm 0.50$&$\pm 0.23$&$\pm 0.35$&$\pm 0.32$&$\pm 0.32$&$\pm 0.33$\\
    \midrule
         R+O30 &17.59&17.14&17.32&17.15&16.48&16.42&\textbf{18.44}&17.07&16.40&\textbf{18.51}&\textbf{18.41}&16.90&17.51\\
&$\pm 0.32$&$\pm 0.44$&$\pm 0.36$&$\pm 0.36$&$\pm 0.60$&$\pm 0.44$&$\pm 0.21$&$\pm 0.32$&$\pm 0.50$&$\pm 0.25$&$\pm 0.26$&$\pm 0.79$&$\pm 0.88$\\
\midrule
         R+M &17.90&17.70&17.58&17.82&17.54&17.73&\textbf{18.53}&17.89&17.48&18.29&\textbf{18.57}&16.18&\textbf{18.80}\\
&$\pm 0.29$&$\pm 0.33$&$\pm 0.28$&$\pm 0.35$&$\pm 0.34$&$\pm 0.29$&$\pm 0.31$&$\pm 0.31$&$\pm 0.43$&$\pm 0.39$&$\pm 0.31$&$\pm 1.86$&$\pm 0.25$\\
\midrule
R+T&0.35&\textbf{18.51}&15.95&\textbf{18.53}&\multicolumn{2}{c}{Unavailable}&17.97&18.34&17.79&18.19&\textbf{18.49}&14.94&-2.26\\
&$\pm 2.34$&$\pm 0.26$&$\pm 1.29$&$\pm 0.34$&&&$\pm 0.35$&$\pm 0.31$&$\pm 0.30$&$\pm 0.41$&$\pm 0.30$&$\pm 2.72$&$\pm 0.44$\\
    \midrule
         HC+O &6062&6759&8255&8475& - & - &6213&9165&8736&\textbf{9448}&\textbf{9973}&\textbf{9646}&7904\\
&$\pm 405$&$\pm 352$&$\pm 474$&$\pm 426$& & &$\pm 216$&$\pm 212$&$\pm 315$&$\pm 154$&$\pm 223$&$\pm 265$&$\pm 323$\\
\midrule
HC+MA&9668&9853&9780&10003&\multicolumn{2}{c}{Unavailable}&7166&9898&9515&\textbf{10079}&\textbf{10318}&\textbf{10235}&8369\\
&$\pm 282$&$\pm 367$&$\pm 403$&$\pm 394$& & & $\pm 321$&$\pm 134$&$\pm 170$&$\pm 443$&$\pm 328$&$\pm 202$&$\pm 324$\\
 \midrule
         H+M &3151&3258&\textbf{3354}&3295& - & - & \textbf{3309}&3265&\textbf{3305} &3253&3278&3180&3190\\
&$\pm 103$&$\pm 55$&$\pm 10$&$\pm 35$& & & $\pm 20$&$\pm 19$&$\pm 16$&$\pm 60$&$\pm 23$&$\pm 54$&$\pm 58$\\
\midrule
H+O+S&987&2245&3179&\textbf{3261}&\multicolumn{2}{c}{Unavailable}&1785&\textbf{3260}&3251&3257&2992&2867&\textbf{3300}\\
&$\pm 78$&$\pm 233$&$\pm 30$&$\pm 34$&&&$\pm 316$&$\pm 26$&$\pm 26$&$\pm 19$&$\pm 300$&$\pm 290$&$\pm 21$\\
    \midrule
W+M &1479&\textbf{4694}&4460&\textbf{4898}& - & - &4454&4071&4341&4446&\textbf{4595}&4268&4319\\
&$\pm 280$&$\pm 169$&$\pm 84$& $\pm 155$& & & $\pm 51$&$\pm 190$&$\pm 146$&$\pm 131$&$\pm 147$&$\pm 158$&$\pm 107$\\

\midrule
W+O+S&856&931&3959&\textbf{4102}&\multicolumn{2}{c}{Unavailable}&3408&3377&3467&3828&\textbf{4127}&\textbf{4105}&3961\\
&$\pm 70$&$\pm 113$&$\pm 77$&$\pm 100$&&&$\pm 300$&$\pm 37$&$\pm 65$&$\pm 87$&$\pm 79$&$\pm 76$&$\pm 96$\\
\midrule
\midrule
\multicolumn{13}{l}{(batch size = 16)} \\
         HC+O  &4108&5303&5826&6787&7053&7306&5984&7245&7387&\textbf{8059}&7071&\textbf{7748}&\textbf{7459}\\
&$\pm 399$&$\pm 115$&$\pm 366$&$\pm 378$&$\pm 239$&$\pm 362$&$\pm 319$&$\pm 191$&$\pm 204$&$\pm 259$&$\pm 762$&$\pm 245$&$\pm 169$\\
\midrule
         H+M &3166&3260&3193&3247&3154&\textbf{3290}&3041&\textbf{3262}&3221&\textbf{3303}&3211&3249&3243\\
&$\pm 70$&$\pm 28$&$\pm 61$&$\pm 62$&$\pm 122$&$\pm 33$&$\pm 52$&$\pm 31$&$\pm 49$&$\pm 19$&$\pm 58$&$\pm 34$&$\pm 52$\\
\midrule
W+M&2138&3021&3409&\textbf{3767}&2837&3292&3644&3483&\textbf{4013}&3483&3600&\textbf{3672}&3431\\
&$\pm 251$&$\pm 176$&$\pm 267$&$\pm 246$&$\pm 431$&$\pm 506$&$\pm 128$&$\pm 117$&$\pm 137$&$\pm 182$&$\pm 135$&$\pm 106$&$\pm 182$\\
    \bottomrule[1pt]
    \end{tabularx}
    \caption{Average Rewards.}
    \label{tab:rewards}
\vspace{-5pt}
\end{table*}